\documentclass{article}

\usepackage{graphicx}
\usepackage{tabularx}
\usepackage{multicol,multirow}
\usepackage{amsmath,amssymb,amsfonts}
\usepackage{mathrsfs}
\usepackage{amsthm}
\usepackage{apacite}
\usepackage{rotating}
\usepackage{appendix}
\usepackage[authoryear]{natbib}
\usepackage{ifpdf}
\usepackage[T1]{fontenc}%
\usepackage{times}%
\usepackage{textcomp}%
\usepackage{xcolor}
\usepackage{hyperref}
\usepackage{authblk}
\usepackage{booktabs}

\usepackage{bigstrut}
\usepackage{tikz}
\usepackage{MnSymbol}
\usepackage{soul}
\usepackage{pifont}

\usepackage[margin=1in]{geometry}

\newcommand{\Cross}{\mathbin{\tikz [x=1.4ex,y=1.4ex,line width=.2ex] \draw (0,0) -- (1,1) (0,1) -- (1,0);}}

\title{Explainable Machine Learning for Public Policy: Use Cases, Gaps, and Research Directions}
\date{}
\author[]{Kasun Amarasinghe}
\author[]{Kit T. Rodolfa}
\author[]{Hemank Lamba}
\author[]{Rayid Ghani}
\affil[]{Machine Learning Department and Heinz College of Information Systems and Public Policy, Carnegie Mellon University}

\begin{document}
\maketitle

\abstract{Explainability is highly-desired in Machine Learning (ML) systems supporting high-stakes policy decisions in areas such as health, criminal justice, education, and employment.  
While the field of explainable ML has expanded in recent years, much of this work has not taken real-world needs into account. 
A majority of proposed methods are designed with \textit{generic} explainability goals without well-defined use-cases or intended end-users and evaluated on simplified tasks, benchmark problems/datasets, or with proxy users (e.g., AMT). 
We argue that these simplified evaluation settings do not capture the nuances and complexities of real-world applications. As a result, the applicability and effectiveness of this large body of theoretical and methodological work in real-world applications are unclear.
In this work, we take steps toward addressing this gap for the domain of public policy. First, we identify the primary use-cases of explainable ML within public policy problems. For each use case, we define the end-users of explanations and the specific goals the explanations have to fulfill. 
Finally, we map existing work in explainable ML to these use-cases, identify gaps in established capabilities, and propose research directions to fill those gaps to have a practical societal impact through ML. The contribution is 1) a methodology for explainable ML researchers to identify use cases and develop methods targeted at them and 2) using that methodology for the domain of public policy and giving an example for the researchers on developing explainable ML methods that result in real-world impact.}

\paragraph{Policy Statement:}
Despite a rich body of methodological work in explainable ML, little guidance exists for building systems that meet the needs of actual policy applications. This paper seeks to fill that void by mapping out explainability use-cases in public policy settings and comparing the capabilities of existing methods against the requirements of each use-case's stakeholders.
We believe this work serves public policy in two ways: (1) for researchers, a call for empirical, application-focused development and evaluation of explainability methods that will lead to systems better suited to provide social impact; and (2) for policymakers and ML practitioners, a guide to navigating the complex landscape of ML explainability when designing and evaluating applied ML systems that support their policy objectives.

\section{Introduction}

Machine Learning (ML) systems are increasingly supporting high-stakes public policy decisions in areas such as criminal justice, education, healthcare, and social services \cite{Potash2020, Ye2019, Rodolfa2020, Bauman2018, Caruana2015}.
As users of these systems have grown beyond ML experts and the research community, the need to better interpret and understand them has grown as well, particularly in the context of high-stakes decisions that affect individuals' health or well-being \cite{Lakkaraju2016, Rudin2019, Lipton2018}.
Likewise, new legal frameworks reflecting these needs are beginning to emerge, such as the \textit{right to explanation} in the European Union's General Data Protection Regulation \cite{Goodman2017}.

Against this background, research into  \textit{explainability/ interpretability}\footnote{We combine the two terms \textit{interpretability} and \textit{explainability} and use both terms to refer to the ability to understand, interpret, and explain ML models and their predictions}
of ML models has experienced rapid expansion and innovation in recent years, with a focus on method development.
A range of methods have been developed that broadly fall into two categories: 1) inherently interpretable models \cite{Rudin2019, Ustun2013, Lakkaraju2016, Caruana2015, Yang2017}, and 2) post-hoc methods for explaining (opaque) complex models and/or their predictions \cite{Ribeiro2016, Ribeiro2018, Lundberg2017, Lundberg2018a, Lundberg2018, Bach2015, Mothilal2020, wachter2018counterfactual}.
While this expansion of the field has yielded a rich body of methodological work, recently, the community has begun to highlight the shortfalls such as the lack of consistent language and definitions, the lack of clearly defined explainability goals, and desiderata; and the lack of consensus on metrics and methods of evaluating the quality of explanations \cite{Lipton2018, Doshivelez2017, Weller2019, Bhatt2020, Sokol2020, Hase2020, Buccina2020, Chen2021, Bhatt2020a}. 
In addition to the critique above, we argue that there are two key areas where most existing work related to explainable ML methods fall short:
\begin{enumerate}
\item Explainability methods are often developed as ``general-purpose'' methods with a broad and loosely-defined goal,  such as perceived transparency, and not to address specific needs of real-world use-cases. 
\item Explainability methods are not rigorously evaluated to adequately reflect their effectiveness in real-world settings. Barring a few exceptions \cite{Ustun2019a, Caruana2015, Lundberg2018, Jesus2021}, much of the existing work is designed and developed for benchmark classification problems, often with synthetic data and usually validated with user studies limited to users in research settings such as Amazon Mechanical Turk  \cite{Ribeiro2016, Lundberg2017, Bach2015, Hu2019, Plumb2018, Zeiler2014, Simonyan2013}.
\end{enumerate}
The result is a body of methodological work without clearly identified use-cases and more importantly, without established real-world utility, making it difficult for practitioners to select and deploy these methods with any confidence.
A necessary first step for filling these gaps is clearly defining how explainable ML fits into a decision-making process.
As explainability is not a monolithic concept and can play different roles in different applications \cite{Molnar2019, Lipton2018}, this process requires extensive domain/application-specific efforts.

In this paper, we use our experience working with government agencies and non-profits and focus on applications of ML to public policy problems.
Among the broad range of intervention points that the domain of policy presents to ML (e.g., policy design, evaluation, administration), we focus our attention on policy administration tasks where predictive ML models are used to support human decisions with objectives of improving the efficiency of resource usage, the effectiveness of interventions, and equity of outcomes.     
We seek to define the role of explainable ML in these domains and how we can use it to improve policy and social outcomes.
To that end, this paper has the following contributions:
\begin{enumerate}
    \item Identifying the primary use-cases of ML explanations in public policy applications.
    \item For each use case, identifying the goals of the explanation methods, the end-users, and the explanation needs.
    \item Identifying research gaps by comparing the known capabilities of the existing body of work to the needs of the use-cases
    \item Proposing research directions to develop effective explainable ML systems that would be targeted for the needs of real-world use cases and lead to improved policy decisions and consequently improved societal outcomes 
\end{enumerate}

The primary goal of this work is to bridge the gap between methodological research in explainable ML and the needs of policy applications. We believe that effective human-ML collaborative decision-making can profoundly impact policy decision-making processes, and the explainability of ML systems plays a critical role in human-ML collaboration. Thus, bridging the gap between explainable ML methods and applications is paramount.

As computer scientists who develop and apply ML algorithms to improve policy decision making processes, this paper is our attempt at connecting the ML research community with problems in public policy where explainable ML can impact consequential decisions. 
It is worth noting that there have been other pieces in the literature that are similarly motivated in understanding how we can use explainable ML in practical applications \cite{Bhatt2020a, Belle2021, Hong2020}. 
These efforts have primarily focused on identifying how the ML community is using existing interpretability methods. 
For instance \cite{Bhatt2020, Bhatt2020a, Hong2020} conduct semi-structured interviews with different stakeholders in industry to understand how they incorporate interpretable ML in their workflows and gain valuable insights on how ML practitioners perceive the methods of explainable ML. 
We believe our work supplements said pieces through an in-depth analysis of a single domain where we look at potential uses, the needs, where there are gaps in current research in meeting those needs, and how the practitioners and researchers can work together in bridging those gaps. 
We do not intend this work to be a thorough survey of existing work in explainable ML (since there are already excellent articles on that topic \cite{Arya2019, Adadi2018, Molnar2019, Bhatt2020, Guidotti2018}) but rather to highlight the needs of the domain, map the capabilities of existing approaches to those needs, identify gaps, and propose concrete steps to bridge those gaps. 
The primary audience of this work is the ML research community that designs and develops explainable ML systems that may be implemented in public policy decision-making systems. 
We believe that this discussion will serve as a framework for designing explainable ML methods and evaluation setups with an understanding of the following: 
\begin{enumerate}
    \item The purpose the explanations serve and the related policy/societal outcome
    \item The end-user of the explanations, the exact decisions they would make based on the explanations, and the intended impact of explanations on their decisions
    \item How to measure the effectiveness of generated explanations in helping end-users make better decisions that result in improved public outcomes (e.g., metrics that reflect decision outcomes)
\end{enumerate}
Furthermore, we believe this work could serve as a guide for policy practitioners who procure and embed ML systems in their decision processes to perform more informed evaluations of the explainable ML systems they procure.  

Although the focus on public policy applications reflects the area of expertise of the authors and a domain that is beginning to use ML tools for assisting many high-stakes decisions, we believe our approach to defining the role of explainable ML in this setting will be valuable as a template for other domains as well.

\section{Use of Machine Learning in Public Policy Problems}
\label{sec:policy_characteristics}
Machine learning (ML) models can analyze large amounts of data to identify patterns and make predictions about future events (e.g., the risk of an evicted individual ending up homeless in the next year, the risk of a student not graduating high-school on time, processing legislative bills to understand the policy areas covered in the bill). 
These predictions can provide data-driven insights to supplement human expertise to inform decision-making and the policy domain presents a range of such decision points.  
However, it is important to note that policy decision making is a complex, human, and political process, and there are many challenges to using ML in this domain.
For instance, policymaking typically involves trade-offs between competing social values \cite{Parkhurst2016, Saltelli2017}, whereas ML algorithms require explicitly defined objectives and weighing competing objectives \cite{Coyle2020}. 
Additionally, while ML predictions can provide useful information policy decisions are typically not solely based on technical evidence \cite{Parkhurst2016} and that the power of persuasion and in some cases manipulation, plays a critical role in legislative processes \cite{Cairney2017, Zahariadis2003}.

In this work, we focus on policy administration decisions where ML predictions assist resource allocation and intervention decisions at a highly granular level (e.g., predicting the risk of future mental health crises to do individual-level proactive mental health outreach). 
Typically these systems are designed to improve the efficiency of resource utilization, intervention effectiveness, and equity of outcomes. 
To mitigate the ambiguity and uncertainty inherent to policy processes, we assume a continuous partnership between partnering policy practitioners and ML practitioners in defining the goals, parameters of operationalizing ML predictions, how to measure success, possible risks, and mitigation strategies (e.g., bias and equity).
In this work, we draw on our experience in partnering with governments to develop and implement human-ML collaborative policy administrative systems where ML predictions (and potential ML explanations) supplement decision-makers' domain expertise. 

To illustrate the applicability of ML to policy administration settings, we focus on the common task of early warning systems (EWS) that are prevalent in different policy domains. 
In an EWS, the ML model is used to identify entities (e.g., people, schools, buildings, locations, etc.) for some intervention, based on a predicted risk of some (often adverse) outcome, such as an individual getting diagnosed with a disease in the next year, a student not graduating high school or college on time, a tenant getting harassed by their landlord, or for a child getting lead poisoning within the next year \cite{Bauman2018, Ye2019, Rodolfa2020}. 
While there are several other policy problem templates that ML is used for, such as inspection targeting, scheduling, routing, and policy evaluation, we use early warning systems to illustrate our ideas in this paper.

\subsection{Characteristics of ML applications in public policy}
Several characteristics of typical public policy problems set them apart from standard benchmark ML problems and data sets often used to evaluate newly proposed algorithms:

\textbf{Non-stationary environments.}
In a policy context, ML models use data about historical events to predict the likelihood of either the occurrence of an event in the future or the existence of a present need, and the context around the problem changes over time.
This non-stationary nature in the data introduces strong temporal dependencies that should be considered throughout the modeling pipeline and makes these models susceptible to errors such as data leakage \cite{Samala2020, Kaufman2011}. 
For instance, the use of standard randomized \textit{k}-fold cross-validation as a model selection strategy can create training sets with information from the future, which would not have been available at model training time.  

\textbf{Evaluation metrics reflect real-world resource constraints.}
The mental health outreach in \cite{Bauman2018} was limited by staffing capacity to intervene on only 200 individuals at a time, and the rental inspections team in \cite{Ye2019} could only inspect around 300 buildings per month.
Resource constraints such as these are inherent in policy contexts, and the metrics used to evaluate and select models should reflect the deployment context. 
As such, these applications fall into the \textit{top-k} setting, where the task involves selecting exactly $ k $ instances as the ``positive'' class \cite{Liu2016}.  
In such a setting, we are concerned with selecting models that work well for precision in the top $k$\% of predicted scores \cite{Boyd2012} rather than optimizing accuracy or AUC-ROC (as often done in ``standard'' classification problems), which would be sub-optimal. 

\textbf{Heterogeneous data sources with strong spatiotemporal patterns.}
Developing a feature set that adequately represents individuals in policy applications typically entails combining several heterogeneous data sources, often introducing complex correlation structures to the feature space not usually encountered in ML problems used in research settings.
For instance, in \cite{Bauman2018}, the ML model combines data sources such as criminal justice data (jail bookings), emergency medical services data (ambulance dispatches), and mental health data (electronic case files) to gain a meaningful picture of an individual's state.
Additionally, temporal patterns in the data are often particularly instructive, requiring further expansion of the feature space to capture the variability of features across time (number of jail bookings in the last six months, 12 months, and five years).
The combination of such features across a range of domains, geographies, and time frames yields a large (and densely-populated) feature space compared to typical structural data-based ML problems we encounter in research settings. 

\subsection{Socio-technical systems}
 Typical ML-supported public policy decision-making systems have at least four types of users that interact with ML models at different stages of the process:
 
\begin{enumerate}
\item \textbf{ML practitioners} who build the ML components of the system. 
\item \textbf{High-level decision-makers/regulators} who determine whether to adopt the ML models in their decision-making processes or are responsible for auditing the ML models to ensure intended policy outcomes.
\item \textbf{Action-takers} (e.g., social workers, health workers, employment counselors) who act and intervene based on the recommendation of the ML model. 
Most policy applications of ML do not involve fully automated decision-making, but rather a combined system of ML model and action-taker that we consider as one decision-making entity.
Action-takers often make two types of decisions: deciding whether to accept/override the model prediction for a given entity \textit{(whether to intervene)} and deciding which intervention to select in each case \textit{(how to intervene)}.
\item \textbf{Affected individuals} that are impacted by the decisions made by the combined human-ML system. 
\end{enumerate}

\begin{table*}[t!]
    \caption{Use-cases of explainable ML in public policy applications}
    \centering
    \vspace{0.5em}
    \begin{tabular}{p{3cm}  p{4cm}  p{6cm}}
      \textbf{Use case}  & \textbf{End users} & \textbf{Intended use of explanations} \\ 
      \midrule 
      Model debugging & ML System Developers & Uncover and fix errors in the ML pipeline such as leakage or biases
      \\ 
      \midrule
      Trust \& adoption & Policymakers, Regulators, \& Action Takers & Help users understand how the model makes decisions, evaluate its reasonableness, and sufficiently trust the model for adoption
      \\  
      \midrule
      Whether to intervene & Action Takers &  Improve the decision making system performance by helping action takers identify correct and unreliable predictions by explaining how the model arrived at individual risk scores
      \\ 
      \midrule
      How to intervene & Action Takers & Improve intervention choices by helping action takers understand factors that contribute to risk 
      \\ 
      \midrule
      Recourse & Affected Individuals & Help affected individuals take action to improve their outcomes in the future or appeal decisions based on inaccurate data\\ 
      \bottomrule
    \end{tabular}
    \label{tab:usecases}
\end{table*}

\section{The Role of Explainable ML in Public Policy Applications}
\label{sec:usecases}
Based on our extensive experience working on over a hundred such projects in collaboration with governments and non-profits and through extensive discussions with stakeholders in public policy settings including policymakers, directors of agencies, policy analysts, end-users such as counselors and social workers, as well as the the public that is impacted, we identify five primary use-cases for explainable ML in a public policy decision-making process (see Table \ref{tab:usecases}). 
For each use-case, we identify the end-user(s) of the explanations, the goal of the explanations, and the desired characteristics of the explanations to reach that goal for that user.
To better illustrate the use-cases, we will make use of concrete applications drawn from our work---preventing adverse interactions between police and the public (citation omitted due to blind review)---to serve as a running example. 
Many applied ML contexts share a similar structure, such as: supporting child welfare screening decisions \cite{Chouldechova2018}, allocating mental health interventions to reduce recidivism \cite{Rodolfa2020, Bauman2018}, intervening in hospital environments to reduce future complications or readmission \cite{Ramachandran2020}, recommending training programs to reduce risk of long term unemployment \cite{Zejnilovic2020}. 

\textbf{Illustrative example:}
Adverse incidents between the public and police officers, such as unjustified use of force or misconduct, can result in deadly harm to citizens, decaying trust in police, and less safety in affected communities.
To proactively identify officers at risk for involvement in adverse incidents and prioritize preventative interventions (e.g., counseling, training, adjustments to duties), many police departments make use of Early Intervention Systems (EIS), including several ML-based systems (see \cite{Carton2016} for an example).
The prediction task of the EIS is to identify $k$ currently active officers who are most likely to be involve in an adverse incident in a given period in the future (in the next 12 months), where the intervention capacity of the police department determines $k$.
The EIS uses a combination of data sources such as officer dispatch events; citizen reports of crimes; citations, traffic stops, and arrests; and employee records to represent individual officers and generates labels using their history of adverse incidents \cite{Carton2016}.

\subsection{Use Case 1: Model debugging}
ML model-building workflows are inherently iterative, and one critical piece of this workflow is the continuous feedback provided by sanity checks on the model(s) to see if they \textit{make sense} and free from errors.
A primary goal of explanations at this early stage is to help the system developers identify and correct errors in the models.
Common errors such as data leakage (the model having access to information at training/building time that it would not have at test/deployment/prediction time) \cite{Kaufman2011}, and spurious correlations/biases (that exist in training data but do not reflect the deployment context of the model) are often found by observing model explanations and finding predictors that should not show up as highly predictive \cite{Ribeiro2016, Caruana2015}. 
ML models trained to predict/detect real-world events typically learn from messy data that capture only a partial view of individuals or entities and are highly susceptible to surfacing spurious correlations.  
Therefore, having additional insight into what the ML model is learning and how it makes decisions through explainable ML can support the model evaluation process. 
For instance, in \cite{Caruana2015}, the authors elaborate on how explanations helped surface such errors in a model trained to identify pneumonia patients with high mortality risk. The model explanations showed that the model assigned low-risk scores to asthma patients because the model did not have access to the information that asthma patients routinely received a more intensive care regimen.  

\textit{E.g.}
In the EIS, an adverse incident gets determined to be \textit{unjustified} a long time after the incident date. 
When training an ML model with the entire incident record, accidentally using the future \textit{determination state} of the incident can introduce data leakage. 
In this case, explanations could uncover that a feature such as the \textit{case disposition code} is considered important by the model when it takes a value related to the determination state and can point the ML practitioner or a domain expert to recognize that information has leaked from the future.  

\subsection{Use Case 2: Building trust for model adoption}
Decision-makers have to sufficiently trust the ML model to adopt and use them in their processes. 
Trust, in general, is a common motivating theme cited by explainable ML work  \cite{Ribeiro2016, Lipton2018, Lundberg2018a}. 
Furthermore, there has been significant emphasis laid on developing `trustworthy ML' systems of which explainable ML is considered to be a key element \cite{li2022}.
In our experience, tust in human-ML collaboration takes two forms in policy contexts: 
1) high-level decision-maker's trust in the model that leads to its adoption, 
and 2) action-taker's trust in the model's predictions that leads to individual actions/interventions.
This use case focuses on the former, where the goal of explanations is to help users (policymakers, organizational leadership, etc.) understand and adequately trust the model's overall decision-making process.\footnote{It is important to note that explainability is not the only factor that affects user trust. In a policy context, factors such as 1) stability of predictions, 2) the training users have received, and 3) user involvement in the modeling process, also impacts user trust \cite{Ackermann2018}.} 

The role of the explanation, in this usecase, is to both help the users understand what factors are affecting the model predictions, as well as characteristics of individuals that are being scored as high or low risk. 
Since the user in this instance is not an ML expert but has expertise in the application domain, communicating the explanation in a way that increases the chances of building trust is critical. 
Further

\textit{E.g.}
In the EIS, the explanations should inform the ranking officer at the PD---who acts as the regulator---of the factors that lead to increasing/decreasing a police officer's risk score \cite{Carton2016}. 
In that instance, \textit{"A high number of investigations in the last 15 years"} is an interpretable indicator while \textit{"positive first principal component of the arrest code"} is not.

\subsection{Use Case 3: Deciding Whether to Intervene}
\label{sec:usecase3}
No ML model makes perfect predictions, especially when predicting rare real-world events.  For example, consider an ML model that predicts children and homes at risk of future lead hazards for allocating limited inspection and remediation resources. 
If only 5\% of households have lead hazards, a model that identifies these hazards with a 30\% success rate (precision) would provide a significant improvement over a strategy of performing random inspections, but would still be incorrect 70\% of the time. 
In the ideal case, the action-taker in the loop (the lead inspection team) would use their expertise to determine when to follow and act on the model’s recommendation and when to override it, resulting in an improved list of $ k $ entities.
This use-case is closely related to the notion of trust we discussed in the above use-case, but at the level of individual predictions and with the end-user being the action-taker. 

Effective explanations can help users, combined with their domain expertise, determine when the model is wrong and improve the overall decisions made by the combined Human-ML system.  
Therefore, the goal of explanations in this use case is to help the action-taker decide \textit{whether to intervene} given the model prediction and its explanations such that the performance of the decision-making system improves (e.g., precision@$k$ in the example above). 
As the end-users are domain experts, the user-interpretability requirement from the above use-case holds for the explanations. 
This use-case has been the most commonly studied use case in explainable ML literature, albeit in non-policy settings. 
For instance, in \cite{Ribeiro2016}, the authors study whether explanations generated from their method (LIME) can highlight the predictors that contributed the prediction and assist users to identify `unreasonable' predictions through a simulated-user study, \cite{Lundberg2018} studied the ability of explanations to assist physicians detect hypoxemia risk during surgery, \cite{Jesus2021} studied the ability to assist fraud analysts detect credit card fraud with explanations of ML predictions.

\textit{E.g.}
In the EIS, if an explanation exists for each officer in the top-$k$, outlining the factors contributing to the risk score, the internal affairs division---who decides \textit{whether to intervene}---can use those explanations to determine the reliability of the model's recommendation to act on it or override it. 

\subsection{Use Case 4: Deciding How to Intervene}
While ML models can help identify entities that need intervention, they often provide little to no guidance on selecting interventions. 
For instance, consider a model that predicts students' risk of not graduating high school on time. 
A student might be at risk due to several reasons, such as struggling with a specific course, bullying, transportation issues, health issues, or family obligations.
Each of those reasons would require a different type of assistive intervention. 
ML explanations can highlight the predictors that contributes to the risk score, and could help a teacher or other domain expert identify the reasons behind the predicted high risk of a student.  

Therefore, in this use case, the goal of the explanation is to help the action-taker decide \textit{how to intervene} and often choose among one of many possible interventions available. 
While typical ML explanations are not truly causal, the factors highlighted in the explanation can provide valuable information to a domain expert in choosing interventions. 
While there hasn't been studies on this use-case for policy administration decisions to the best of our knowledge, there have been a few efforts other domains where researchers have investigated using explainable ML fur supporting recommending actions. 
For instance, \cite{afzaal2021} showed that explanations of student performance predictions can be used to recommend actions to students in self-regulated learning, \cite{Albreiki2022} studied how explanations from ML can be used to recommend remedial actions to low performing students with the goal of improving learning outcomes, and \cite{Sajja2021} demonstrated the use of explainable ML predictions of consumer behavior in helping fashion designers plan for new products. 



\textit{E.g.}
Consider an officer flagged by the EIS, with an explanation indicating that the model is prioritizing features related to the type of dispatches the officer was assigned to in the last few months. 
Upon further inspection of the data, it can be seen that the officer had been dispatched to high-stress situations regularly.
In this instance, a possible intervention is reassigning duties or putting them on low-stress dispatches after a series of high-stress dispatches.

\subsection{Use Case 5: Recourse}
When individuals are negatively impacted by ML-aided decisions, providing them with a concrete set of actionable changes that would lead to a different decision is critical. 
This ability of an individual to affect model outcomes through actionable changes is called recourse \cite{Ustun2019}. 
While recourse has been studied independently from explainable ML \cite{Ustun2019, Konig2021}, ML explanations have the potential to help individuals seek recourse in public policy applications \cite{Karimi2020a, Karimi2020, wachter2018counterfactual}.  

In this use-case, there are two explanation goals: 1) helping the user understand the reasons behind the current decision, allowing them to discover any inaccuracies in the model and /or data and dispute the decision, and 2) helping the user identify the set of actionable changes that would lead to an improved decision in the future.  
As the user in this use-case is the affected individual, the explanations that indicate the reasons behind the decisions should be mapped to a domain that is understandable by the individual. 
Furthermore, the explanations should recommend feasible and actionable changes (e.g. reducing age by 10 years vs reducing debt).

\textit{E.g.} 
In the EIS, the affected individual is the flagged  officer.
If the officer is provided with explanations indicating the reasons behind the elevated risk score and actionable changes that could reduce their risk score, they could either point any inaccuracies or take measures themselves (in addition to the intervention by the PD) to reduce the risk score. 

\section{Current State of Explainable ML}
\label{sec:current_state}
In this section, we summarize the existing approaches in explainable ML.
It is worth noting that the intention here is not to provide an in-depth and comprehensive literature review but rather a broad view of existing approaches and discuss how they apply to the public policy settings described above. 
We refer readers to \cite{Arya2019, Adadi2018, Molnar2019, Bhatt2020, Guidotti2018} for more comprehensive reviews of existing work.

\subsection{Existing work in explainable/interpretable ML}
Existing approaches broadly fall into two categories: 1) inherently interpretable ML models, and 2) post-hoc methods for explaining opaque ML models.\footnote{Note that model opacity can be a reflection of either 1) the model being too complex to be comprehensible, or 2) the model being proprietary \cite{Rudin2019}. In this paper, we focus on opacity created through model complexity.}
ML explanations take two forms: 1) explaining individual predictions (local explanation), and 2) explaining the overall behavior of the models (global explanation).
Typically, local explanations are intended to help users understand \textit{why} the model arrived at the given prediction for a given instance, while global explanations explain \textit{how} the model generally behaves \cite{Plumb2018}. 
Table \ref{tab:method_categories} summarizes the existing approaches.

\begin{table*}[t!]
    \centering
    \caption{A summary of existing approaches for explainable ML}
    \vspace{0.5em}
    \begin{tabular}{p{4.3cm} | p{1.2cm} | p{1.2cm} | p{1.2cm} | p{1.2cm} | p{1.7cm} }
    \multirow{3}{*}{Approach} 
    & \multicolumn{4}{c |}{\centering{Post-hoc methods}} 
    & \multirow{3}{=}{\centering{Interpretable Models}}\\ 
    & \multicolumn{2}{c |}{Local} &  \multicolumn{2}{c |}{Global} & \\ 
    & \centering{Model agnostic} & \centering {Model specific} & \centering{Model agnostic} & \centering{Model specific} & \\
    \hline 
    \bigstrut Sparse models\footnotemark &  &  &  &  & $\checkmark$ \\
    \hline
    \bigstrut Decision Rules \& Sets\footnotemark  &  &  &  &  & $\checkmark$ \\
    \hline
    \bigstrut Gen. additive models\footnotemark  &  &  &  &  & $\checkmark$ \\
    \hline
    \bigstrut Local surrogate models\footnotemark & $\checkmark$ & $\checkmark$ &  &  & \\
    \hline
    \bigstrut Permutation (Shapley values)\footnotemark & $\checkmark$ & $\checkmark$ &  & $\checkmark$ & \\
    \hline
    \bigstrut Global rule extraction\footnotemark &  & $\checkmark$ & $\checkmark$ &  & \\
    \hline
    \bigstrut Gradient-propagation\footnotemark &  & $\checkmark$ &  &  & \\
    \hline
    \bigstrut Counterfactuals\footnotemark & $\checkmark$ & $\checkmark$ &  &  & \\
    \hline
    \bigstrut Example based\footnotemark & $\checkmark$  &   & $\checkmark$  & & \\ 
    \hline
    \end{tabular}
    \label{tab:method_categories}
\end{table*}
\footnotetext[4]{\cite{Ustun2013, Ustun2019a, Hu2019, Zeng2017}}
\footnotetext[5]{\cite{Lakkaraju2016, Ye2019, Letham2015}}
\footnotetext[6]{\cite{Caruana2015, Lou2012, Hastie1990}}
\footnotetext[7]{\cite{Ribeiro2016, Plumb2018}}
\footnotetext[8]{\cite{Lundberg2017, Lundberg2018a, Lundberg2018, Lundberg2020}}
\footnotetext[9]{\cite{Ribeiro2018, Tsukimoto2000}}
\footnotetext[10]{\cite{Bach2015, Simonyan2013, Baehrens2010, Zeiler2014}}
\footnotetext[11]{\cite{Ustun2019a, Poyiadzi2020, Mothilal2020, wachter2018counterfactual}}
\footnotetext[12]{\cite{Koh2017, Kim2016a}}

\subsubsection{\textbf{Inherently interpretable ML models}}
Inherently interpretable ML models are designed such that an end-user could understand its decision-making process \cite{Lakkaraju2016, Rudin2019}.
In a policy context, an interpretable model could allow a user to (a) understand how the model calculates a risk score (global explanation) and (b) understand what factors contributed to the predicted risk score (local explanation) for a given instance.
Several efforts have focused on developing interpretable models for policy domains, such as those for healthcare and criminal justice \cite{Zeng2017, Caruana2015}. 
These include sparse linear models \cite{Ustun2013, Ustun2019}, sparse decision trees \cite{Hu2019}, generalized additive models \cite{Hastie1990, Lou2013, Lou2012, Caruana2015}, and interpretable decision sets \cite{Lakkaraju2016}.  

Interpretable models often rely on carefully curated representations of data with meaningful input features \cite{Rudin2019}, often through discretization or binary encoding \cite{Ustun2019, Caruana2015, Lakkaraju2016}. 
Distilling complex data spaces into a handful of optimally discretized and meaningful features can entail extensive effort and optimization of its own.
While careful feature preparation is indispensable in any ML application, regardless of the employed ML algorithm complexity, distilling complex and heterogeneous feature spaces typically found in policy settings into a handful of simple features can prove to be particularly challenging. 


\subsubsection{\textbf{Post-hoc methods for explaining black-box ML models}}
Post-hoc methods derive explanations from already trained black-box/opaque ML models. 
As post-hoc methods do not interfere with the model's training process, they enable the use of complex ML models to achieve explainability without the risk of sacrificing performance.
However, as black-box ML models are often too complex to be explained entirely, post-hoc methods typically derive an approximate explanation \cite{Rudin2019, Gilpin2018}, which makes ensuring the fidelity of the explanations to the model a key challenge in this work.
Unlike inherently interpretable models, local and global explanations for opaque complex ML models require different methods.
For both types of explanations, both model-specific and model-agnostic methods exist in literature.    

\vspace{1ex}
\noindent \textbf{Post-hoc local explanations:}
A local explanation in a typical public policy problem is used to understand which factors affected the predicted risk score for an individual entity. 
The most common format of local explanation is feature attribution--also known as feature importance or saliency---where each input feature is assigned an importance score that quantifies its contribution to the model prediction \cite{Baehrens2010, Bhatt2020}. 
Several approaches exist for deriving feature importance scores such as: fitting an interpretable surrogate model (linear classifier) around a local neighborhood of the instance in question \cite{Ribeiro2016, Plumb2018}; feature perturbation based methods for approximating each feature's importance using game-theoretic shapely values \cite{Lundberg2017, Lundberg2018a}; and gradient-based techniques \cite{Zeiler2014, Simonyan2013, Bach2015}. 
Among these approaches, methods such as LIME \cite{Ribeiro2016}, SHAP \cite{Lundberg2017}, and SA \cite{Zeiler2014} are model-agnostic methods, whereas LRP \cite{Bach2015}, Deconvolution \cite{Simonyan2013}, and TreeSHAP \cite{Lundberg2018} are model specific methods. 
MAPLE \cite{Plumb2018} stands out among these methods as it can act both as an inherently interpretable model as well as a model-specific post-hoc local explainer.

Other approaches such as influence functions \cite{Koh2017} as well as prototypes \& criticisms \cite{Kim2016a} make use of data instances, rather than features, to provide local explanations. 
A special form of example-based explanation is counterfactual explanations, which seek to answer the following question:  \textit{"what’s the smallest change in data that would result in a different model outcome?"} \cite{wachter2018counterfactual, Barocas2020, Karimi2020, Waa2018, Mothilal2020, Molnar2019}.
In a top-$k$ setting, the \textit{change in outcome} can be the inclusion vs. exclusion of the individual from the top-$k$ list.
Counterfactual explanations can provide insight on \textit{how to act} to change the risk score, supplementing the feature attribution methods that explain \textit{why} the model arrived at the risk score.

\vspace{1ex}
\noindent \textbf{Post-hoc global explanations: } 
A global explanation in a typical policy problem would be a summary of factors/patterns that are generally associated with high-risk scores, often expressed as a set of rules \cite{Plumb2018, Ribeiro2018}.
Global explanations should enable the users to accurately predict, sufficiently frequently, how the model would behave in a given instance. 
However, deriving global explanations of models that learn highly complex non-linear decision boundaries is very difficult \cite{Ribeiro2016}. 
As a result, the area of deriving post-hoc global explanations is not as developed as local explanation methods.

Some approaches for  global explanations from black-box ML models include: 1) aggregation of local explanations \cite{Lundberg2020}, 2) global surrogate models \cite{Frosst2017}, and 3) rule extraction from trained models \cite{Tsukimoto2000}. 
A noteworthy contribution to deriving globally faithful explanations is ANCHORS \cite{Ribeiro2018}. 
ANCHORS identifies feature behavior patterns that have high precision and coverage in terms of their contribution to the model predictions of a particular class.   
Methods proposed in \cite{Lundberg2020, Ribeiro2018} are model-agnostic and methods presented in \cite{Frosst2017, Tsukimoto2000} are model-specific.

\subsection{Capabilities of existing explainable ML methods and public policy use-cases}
In this section, we characterize the established capabilities of the existing explainable ML methods classes with respect to the use cases we identified. 
To rank capabilities, we use a three-point scale that is based on the level of evidence that existing method evaluations demonstrate for an individual use case (See Table \ref{tab:method_applicability}).
Highlighting the multi-faceted nature of evaluating explainable ML methods, \cite{Doshivelez2017} called for more rigorous approaches in the field and mapped these evaluation studies into a three-tiered framework: (1) functional-grounded evaluation, where the intrinsic qualities of the explanation are evaluated purely through algorithmic means, e.g., the fidelity of the explanation to the underlying ML model is a commonly used metric in functional-grounded evaluation, (2) human-grounded evaluation, where the utility of the explanations are assessed using proxy users or simplified tasks, e.g., users from Amazon Mechanical Turk (AMT) performing a task such as simulating the ML model's prediction given the data and explanation is a commonly used human-grounded evaluation setup, and (3) application-grounded evaluation, where the utility of explainable ML is tested at helping real users (domain experts) perform a real-world task. 
The three-point scale that we present is primarily based on human-grounded and application-grounded evaluations as we are interested in highlighting the proven utility of explainable ML methods in the identified the use-cases. 
Our goal for this ranking is to highlight where the established capabilities in the field fall short of the needs of the use cases based on our research and our experience implementing and evaluating them. 
We use the three broad method categories --- post-hoc local methods, post-hoc global methods, and inherently interpretable models --- for this ranking and assign a rating to the whole group if at least one method in the group satisfies the requirements. 
We define the three-point scale as follows: 
\begin{description}
    \item $ \bigstar \largestar \largestar $: Methods are potentially applicable for the use case. However, we haven't found any human-grounded or application-grounded studies where any method in the class is directly evaluated on the use-case and shown to be effective.
    \item $ \bigstar \bigstar \largestar $: Some evidence of efficacy in the use-case exists through evaluations on simplified /proxy problems and proxy users (human-grounded evaluations). However, no application-grounded studies exist where the efficacy of any method is empirically validated through a well-designed user study in a real-world setting where real users are performing a real task. 
    
    \item $ \bigstar \bigstar \bigstar $: At least one method in the group is validated with an application-grounded evaluation on the use-case with a well-designed user study, which implements the method on a real task, uses real data, presents explanations to real users of the system, and empirically demonstrates the method's efficacy at improving outcomes of interest.  
    
    \item  $ \Cross $ : Methods in the group are not applicable to the use-case
\end{description}

The discussion below summarizes how existing work maps to each use-case and our assessment of the status of current work with respect to these applications.
It's worth noting that inherently interpretable models are potentially applicable to all the use-cases. 
Therefore, we focus on the post-hoc methods in the summaries below.


\begin{table*}[t!]
\centering
    \caption{Capabilities of existing methods with respect to the public policy use cases. Please note that the footnotes indicate the publications the highest rating received is based on}
    \vspace{0.5em}
    \begin{tabular}{l | c c c }
    \textbf{Use-case} & \textbf{Post-hoc Local} & \textbf{Post-hoc Global} & \textbf{Interpretable Models} \\
    \hline
    \bigstrut Model debugging\footnotemark & 
    $ \bigstar \bigstar \largestar $ & 
    $ \bigstar \bigstar \largestar $ & 
    $ \bigstar \bigstar \largestar $ \\
    \hline
    \bigstrut Trust \& adoption\footnotemark &  
    $ \bigstar \largestar \largestar $ & 
    $ \bigstar \largestar \largestar $ & 
    $ \bigstar \largestar \largestar $ \\
    \hline
    \bigstrut Whether to intervene\footnotemark &  
    $ \bigstar \bigstar \largestar$ & 
    $ \Cross $ & 
    $ \bigstar \largestar \largestar $ \\
    \hline
    \bigstrut How to intervene\footnotemark &  
    $ \bigstar \largestar \largestar $  & 
    $ \Cross $ & 
    $ \bigstar \largestar \largestar $ \\ 
    \hline
    \bigstrut Recourse\footnotemark &  
    $ \bigstar \bigstar \largestar $  & 
    $ \Cross $& 
    $ \Cross $ \\
    \hline
    \end{tabular}
    \label{tab:method_applicability}
\end{table*}

\footnotetext[13]{\cite{Ribeiro2016, Caruana2015}}
\footnotetext[14]{\cite{Jacovi2021, Ribeiro2016, Buccina2020, Hase2020, Ribeiro2018}}
\footnotetext[15]{\cite{Ribeiro2016, Lundberg2017, Lundberg2018a, Jesus2021}}
\footnotetext[16]{\cite{Ribeiro2016, Lundberg2017, Lundberg2018a, Bach2015, Plumb2018, Ustun2019a, Lou2012, Hu2019, Lakkaraju2016}}
\footnotetext[17]{\cite{Mothilal2020, Ustun2019a, Karimi2021, Karimi2020, Karimi2020a}}

\subsubsection{Model debugging:} 
Methods for both local and global post-hoc explanations are potentially useful in this use-case.
Global explanations could help identify errors in overall decision-making patterns (e.g., globally important features can help identify data leakage), and local explanations can help to uncover errors in individual predictions. 

There have been several studies that lend evidence to the utility of explainable ML methods in uncovering errors in models \cite{Ribeiro2016, Caruana2015, adebayo2020, Abid2022}. 
For instance, in \cite{Ribeiro2016} authors demonstrate how LIME explanations could help users identify model errors through a simplified image classification task and a text classification task.
As we mentioned above \cite{Caruana2015} shows how explanations can uncover spurious correlations that the model picks up based on gaps that exist in the training data.

Although some recent work lends evidence for the utility of explanations in discovering model errors \cite{Ribeiro2016, Caruana2015}, the efficacy of these methods is not empirically validated through well-defined user trials in real-world applications. 
For instance, in \cite{Ribeiro2016} Ribeiro et al. demonstrate how LIME explanations could help users identify model errors through a simplified image classification task and a text classification task. While these studies show that users performed better with the availability of explanations, we argue the fact that both classification tasks were simplified by introducing errors to the model, and explanations were presented to users from AMT oversimplifies the task and deviates the experimental setting from real-world applications, rendering the efficacy of the method to be inconclusive.    

\subsubsection{Model trust and adoption:} As with model debugging, both global and local explanation methods are potentially applicable. 
 However, as the end-user is the domain expert, explanations will need to be extended beyond feature attribution while preserving fidelity to what the model has learned.  
 While existing methods discuss user trust as a broad goal, to the best of our knowledge, their ability to help regulators or decision-makers adequately trust ML models is not demonstrated through well-defined evaluations or user trials. 
The experimental work on the notion of trust has relied on subjective, self-reported measures of trust in performing a simplified task \cite{Ribeiro2016, Weitz2019, Buccina2020}. 
However, Jacovi and colleagues \cite{Jacovi2021} in their effort of formalizing the notion of trust in ML, argue that simply asking the user whether they trust the model for a simple task does not evaluate the notion of trust in AI, as the users are not assuming any risk, and they argue that relying on an AI with assumed risk is a prerequisite for trust. 
A proxy task that is often employed in the evaluation of explainable ML is `forward-simulation' \cite{Doshivelez2017, Ribeiro2016, Ribeiro2018}, i.e., a person predicting the ML model's outcome given the input and explanation. This ability to accurately anticipate the model's output is considered a proxy signal of trust \cite{Jacovi2021, Hase2020}. However, despite these initial efforts, to the best of our knowledge, there haven't been experimental efforts that study how existing explainable ML methods impact the notion of trust related to model adoption in decision-making processes, and how it relates to the societal outcomes of interest. 
 
 
 \subsubsection{Improving decision making system performance} Feature attribution-based local explanations are potentially applicable to provide the necessary information to the user. 
 However, feature attribution alone may not be sufficient. 
 Users may need more contextual information such as \textit{How does the instance fit into the training data distribution?} \textit{How does the model behave for similar examples? and what factors did it rely on for those predictions?}
 To that end, there have been some efforts to present visual summaries of explanations to the user \cite{Lundberg2017, Lundberg2018, Ribeiro2018} which could potentially be useful in this use-case.
 Therefore, available local explanation methods do provide a good starting point. 
 
In contrast to the above use-cases, there have been a couple of instances where explainable ML methods were tested using an experimental setting consisting of a real task, real data, and real users. In \cite{Lundberg2018}, Lundeberg et al. studied the ability of an explainable ML system to assist anesthesiologists to detect hypoxemia risk during surgery for proactive intervention. They showed that their system---Prescience---armed with an ML model and SHAP explanations was able to outperform the anesthesiologists in identifying real-time hypoxemia risk. However, their experiment failed to isolate the marginal effect of the explanations by failing to compare the performance of the ML model + explanations to just the ML model. Therefore, while the combined system with predictions and explanations outperformed the domain expert, it was not possible to isolate whether the effects were due to the ML model prediction alone or due to the combined system. In \cite{Jesus2021}, Jesus et al. studied the impact of presenting ML explanations from three local post-hoc explainable ML methods to fraud analysts for assisting fraud detection in credit card transactions. While they organize the experiment to isolate the incremental impact of explanations by running appropriate experiment arms, they make several simplifying assumptions in their experimental design. For instance, they resample the data in their experiment to reflect a 50\% rate, whereas, in the real-world context, fraud is a significantly rarer event. Further, they measure the effectiveness of decisions using confusion matrix-based metrics, assuming that all transactions are of the same value, an assumption that is not valid in a real-world business setting. Therefore, while Jesus et al. has taken steps in the right direction, we argue that their experimental setup still does not reflect the deployment context adequately.
 
 Despite the existence of user studies conducted with real data and real users, the significant limitations of these studies to date mean they fail to provide conclusive evidence that existing explanation methods are effective in assisting humans to make improved decisions in this use case.

 \subsubsection{How to intervene:} As the intervention determinations are often individualized, local explanation methods are potentially applicable for generating the reasons behind the risk score. 
 As with the above use-case, users may need more contextual information to supplement the local explanations such as: \textit{how the instance fits into the training data distribution}, and \textit{intervention history} for \textit{similar}---\textit{w.r.t} data and \textit{w.r.t} explanation---individuals. 
To the best of our knowledge, there is a lack of evidence in the existing body of work on the efficacy of using these local explanation methods for informing intervention selection. 

  \subsubsection{Recourse:} 
Feature attribution-based local explanations are potentially applicable for deriving \textit{reasons} behind the decision, and counterfactual explanations can be useful in explaining how to improve the outcomes. The focus of algorithmic recourse work has been on using counterfactual explanations. 
  As simple counterfactual explanations do not guarantee explanations with actionable changes, there has been a range of approaches proposed for deriving counterfactual explanations that are \textit{diverse}, \textit{sparse}, \textit{plausible}, and \textit{actionable} \cite{Poyiadzi2020, Ustun2019a, Mothilal2020, Karimi2020a, Karimi2020, Upadhyay2021}. Karimi and colleagues provide a survey of methods for algorithmic recourse including in \cite{Karimi2021}. 
  Evaluating methodologies effectively has been a challenge for algorithmic recourse methods and Karimi and colleagues call for better benchmarks \cite{Karimi2021}. 
  Therefore, existing method evaluation largely relies on theoretical guarantees and demonstrations using popular experimental datasets such as Adult-income, German credit lending, and COMPAS. While those data sets have some connection to the real-world, they are not reflective of data sets that most real-world organizations in policy settings have, especially in terms of richness, complexity, and spatiotemporal patterns.
  Studies that empirically validate the efficacy of these methods are still lacking. Since there are no user studies for any class of models, we rate post-hoc local methods with one star. 

\section{Gaps and Proposed Research Directions} 
In this section, we use the mapping between methods and use cases to identify gaps in existing explainable ML research compared to the needs of real-world public policy problems and propose a research agenda to fill those gaps. 
We believe that bridging these research gaps is critical for applying ML to social problems and if we are to safely deploy ML systems that lead to effective policy decisions and have a positive and lasting impact on society.

\subsection{Gap 1: Capabilities of existing methods not adequately evaluated in real-world contexts}
The most pronounced gap between explainable ML research and the policy use cases is the lack of evidence of method efficacy established through rigorous application-grounded evaluation studies (in our review, we failed to find any study that met the criteria to achieve a three-star rating in Table 3).
The most common approach to evaluating explainable ML has been to assess the quality of the explanation (the artifact produced by the method) through functional-grounded evaluations. 
Almost all user studies take a human-grounded approach where non-expert users (e.g., users from AMT or users in research settings) perform simplified/proxy tasks such as `forward simulation'. 


A growing body of work that empirically demonstrates the limitations of the functional- and human-grounded approaches has begun to appear recently. 
\cite{Hase2020} describe an experiment where users were asked to perform the task of `forward simulation' and subjectively evaluate the quality of explanations (measuring a concept of ``simulatability''). 
Although simulatability seems unlikely to reflect real-world use of explanations, it is notable that the authors found essentially no relationship between the human-grounded subjective assessments of explanation quality and how users performed on this task. 
Similarly, \cite{Buccina2020} compared three proposed measures of explainable ML: subjective user assessments, user performance on a proxy task (predicting model scores based on explanations, similar to the study by Hase and Bansal), and performance on a decision-making task (assessing the nutritional content of different plates of food). 
Their results indicated that performance on neither the subjective assessments nor the proxy task generalized to performance on the actual decision-making task, highlighting the risks of relying too heavily on these simplified evaluations.
We argue that functional-grounded and human-grounded evaluations are not sufficient to establish the utility of explainable ML methods in domains such as public policy where ML systems learn from highly complex, heterogeneous, and messy data spaces and assist consequential decisions. 
In this work, we are interested in evaluating explainable ML methods on their ability to improve a societal outcome of interest.
Functional-grounded metrics such as fidelity do not guarantee the usefulness of explanations and designing proxy tasks for human-grounded evaluations that capture the nuances and complexities of public policy problems can be challenging.

\subsubsection{Guidelines for adequate evaluation of explainable ML in policy contexts}
Given the limitations of functional- and human-grounded approaches, evaluation studies of ML explanations in policy contexts should focus on application-grounded approaches. 
\cite{Doshivelez2017} defines an application-grounded evaluation as a study where domain experts perform the intended task. 
We further extend those requirements and argue that an adequate application-grounded evaluation of an explainable ML method in a policy context cannot exist in the absence of four key elements: (1) A real policy task and related metrics, (2) users who perform the task in the real world, (3) real-world data related to the task that captures the complexities and nuances of the task, and (4) a robust inference strategy that allows conclusions on the incremental impact of explanations. Unfortunately, the relatively small number of application-grounded explainable ML studies that incorporated some aspects of practical evaluation have consistently lacked at least one (and in many cases multiple) of the elements necessary to offer conclusive evidence of real-world efficacy.
We elaborate on each of these elements and how existing work violates those requirements below:

\begin{description}
    \item[Defining the task:] The task here entails the decision a user would make, the goals of the decision-making process, and the metrics that evaluate its success. It is imperative to pick tasks that align with a well-defined policy/operational goal and metrics that directly measure the success of the task, going beyond general-purpose metrics such as ROC-AUC, Accuracy, and F1-score.  
    For instance, consider the study conducted by \cite{Jesus2021} on a credit-card fraud detection context. While the authors conduct the study with professional fraud analysts performing fraud detection on real-world credit card transactions, they make a simplifying assumption on selecting metrics and choose decision accuracy and other confusion matrix-based metrics. In the context of e-commerce transactions, we argue that the goal is to maximize revenue/profit, and the metric should factor in the transaction value and the relative costs of the two types of errors. Choosing accuracy as the metric ignores both these nuances and thus violates the first requirement of a study.    
    In \cite{Poursabzisangdeh2021}, authors used real-world data and a large cohort of real users. However, they choose to use the task of real estate valuation, and how it relates to a decision and an outcome of interest and the utility of explanations in achieving that goal was not established.  
    \\
    \item[Recruiting users:] Although their time is often scarce, the domain expert users who act on model outputs must be involved in the evaluation process to ensure it reflects the actual deployment scenario.
    As the interaction between model predictions, explanations, and users' domain expertise will dictate the performance of the system, substituting inexperienced users (for instance, from AMT) provides little insight into how well explanations will perform in practice and the results generated \cite{Lundberg2017, Lundberg2018a, Lou2013, Hu2019} may not correlate with results in actual deployment with real users.
    We argue that even if a study implements explainable ML methods on real-world problems and data that capture the nuances and complexities of the domain, the absence of evaluation with domain experts leads to uncertainty of method efficacy. 
    For instance, \cite{Caruana2015} and \cite{Zeng2017} describe evaluations of their inherently interpretable models on a real problem with a clear goal and real-world data. However, both studies failed to conduct studies with actual users of the system to evaluate the incremental utility of explanations in terms of performance improvement in the task. 
    \\
    \item[Data:] To capture the nuances and characteristics of applying ML to a policy area in practice, the use of data from the problem domain is essential. 
    This is particularly important with evaluating explainable ML methods, as simplified or synthetic data sets might provide an overly-optimistic evaluation of their ability to extract meaningful information. 
    Unfortunately, most of the work in this area fails to meet these criteria, focusing only on benchmark ML problems and data sets (e.g., image classification on MNIST data, newsgroups data) \cite{Ribeiro2016, Bach2015, Ribeiro2018, Shrikumar2017}. While benchmark problems play a crucial role in explainable ML method development and refinement, these problems are far removed from deployment contexts encountered in public policy settings and thus fail to provide convincing evidence of method effectiveness in informing the decisions of domain experts. 
    It is important to note that even in studies that do use data from the problem domain, seemingly trivial simplifying assumptions can violate this requirement. For instance, in \cite{Jesus2021}, the authors adjust the class distribution artificially create a 50-50 distribution to remove the `class imbalance' problem, whereas the actual fraud rate is around 15\%. This simplication can impact the expected prior beliefs of fraud analysts and findings of the study.    
    \\
    \item[Defining the inference strategy:] In addition to setting up problems, data, and users consistent with the deployment context, one aspect where existing studies have faltered is the design of the inference strategy of the experiment. It is essential to design the inference strategy to support conclusions on the incremental impact of the explanations in the context. A robust inference strategy entails evaluating the appropriate hypotheses, appropriate control/treatment groups, sufficient sample sizes to preserve statistical power, and analytical methodologies that capture the uncertainties in data.
    Consider the study conducted by \cite{Lundberg2018} where the authors evaluate an explainable ML system on its ability to support anesthesiologists detect hypoxemia during surgeries. The authors implemented the method on the actual task, recruited domain expert users, and used historical data captured during surgeries. 
    Unfortunately, the study fails to evaluate the incremental impact of the explainable ML method as it does not evaluate the appropriate hypotheses. The authors compare the performance between users making decisions only using data and users making decisions with the help of data, ML prediction, and explanation. The study concludes that users perform better when they have access to ML predictions and explanations compared to only having access to the data. 
    However, the authors fail to compare the performance of users with access to data and ML prediction (no explanations) to the user performance with all three pieces of information. We argue that this is an essential component of an experimental design aimed at evaluating the incremental impact of explainable ML, and this oversight prevents us from attributing that performance increment to the explanations.
\end{description}

As we can see, even the relatively few application-grounded explainable ML studies that have incorporated some aspects of practical evaluation have consistently lacked at least one (and in some cases multiple) of the elements necessary to offer conclusive evidence of real-world efficacy (See Table \ref{tab:app_grounded_studies}).
This gap is particularly acute in the context of public policy problems, given the characteristics (See Section 2) that set them apart from ML problems we encounter in research settings. Therefore, we argue that for implementing explainable ML systems in policy settings, there should be concerted efforts from machine learning and policy practitioners to conduct well-designed application-grounded evaluation studies.

\begin{table*}[t!]

\centering
    \caption{Analyzing the existing handful of application-grounded evaluation studies with respec to the the proposed desiderata. Please note that we include studies that at least satisfies one requirement. }
    \vspace{0.5em}
\begin{tabular}{l | cc | c |c | c}
\multicolumn{1}{c |}{\multirow{2}{*}{\textbf{Study}}} & 
\multicolumn{2}{c|}{\textbf{Real Task}} & 
\multicolumn{1}{c|}{\multirow{2}{*}{\textbf{Real Data}}} & 
\multicolumn{1}{c|}{\multirow{2}{*}{\textbf{Real Users} }} & 
\multicolumn{1}{c}{\multirow{2}{*}{\textbf{Inference}}} \\ 
\multicolumn{1}{c|}{} & 
\multicolumn{1}{l|}{\textbf{Decision}} & 
\multicolumn{1}{l|} {\textbf{Metrics}} & 
\multicolumn{1}{c|}{\textbf{}} & 
\multicolumn{1}{c|}{\textbf{}} & 
\multicolumn{1}{c}{\textbf{}} \\ 
\hline
\bigstrut \cite{Jesus2021}   & $\checkmark$  &  $\Cross$       & $\Cross$  & $\checkmark$  & $\checkmark$  \\
\bigstrut \cite{Lundberg2018}& $\checkmark$  & $\checkmark$  & $\checkmark$  & $\checkmark$ & $\Cross$ \\ 
\bigstrut \cite{Poursabzisangdeh2021}& $\Cross$ & $\Cross$ & $\checkmark$ & $\checkmark$ & $\checkmark$ \\
\bigstrut \cite{Caruana2015}& $\checkmark$ & $\checkmark$ & $\checkmark$ & $\Cross$ & $\Cross$ \\
\bigstrut \cite{Zeng2017}&$\checkmark$ & $\checkmark$ & $\checkmark$ & $\Cross$ & $\Cross$ \\
\hline
\label{tab:app_grounded_studies}
\end{tabular}
\end{table*}

\subsubsection{An example evaluation study design that satisfies the desiderata}
To make the desiderata more concrete, this section provides an overview of an experimental design for evaluating an explainable ML method in a policy context that implements the four necessary elements we presented.  
While this example describes a setting that uses a post-hoc explainable ML method, the same setup could be used to evaluate an inherently explainable model. The policy problem presented here is similar to the one discussed in \cite{Bauman2018}.

\paragraph{The policy problem:} The Mental Health Center (MHC) of a mid-sized suburban county is establishing a program to conduct proactive mental health outreach to individuals at risk of criminal justice involvement due to unmet behavioral health needs. The MHC has the resources to perform outreach and assist about 100 individuals each calendar month. The broad policy goal is to minimize repeated criminal justice involvement among county residents, and the specific goal of the human-ML collaborative system would be to maximize the efficient use of county resources.\footnote{It is worth noting that in a typical project, the efficiency goals are often coupled with an equity goal, but for simplicity we narrow the scope of this example down to focus on efficient resource allocation.}

\paragraph{ML Model:} In order to prioritize individuals for outreach, a ML based predictive model is used. Among the county residents who were released from jail in the last two years, the model predicts the risk of each individual being booked into jail within the next 12 months and maps them into an ordinal scale.

\paragraph{Users: } Mental health clinicians of the Mobile Crisis Response Team (MCRT) will act on the predictions of the ML model by conducting outreach and offering appropriate services to these individuals based on their needs. 

\paragraph{Task \& Metrics:} Given an individual that is deemed to be at risk of future criminal justice involvement (by the ML model), the MCRT clinician is tasked with verifying the risk and selecting them for mental health outreach. In different experimental conditions, the clinicians would have different information pieces at their disposal to make this decision. 
Since the goal is maximizing efficiency, the objective is to correctly identify people who are at actually risk. Therefore, the metric we should be optimizing for should capture how accurate the MCRT clinicians are at identifying the individuals for outreach, i.e., given that an individual is selected for outreach, maximizing the probability that they are actually at risk of being booked into jail the following year, which can be captured by maximizing the positive predictive value (PPV)/Precision.

\paragraph{Data: }  The ML model learns from historical data from the county jail, emergency medical data, and behavioral health service involvement data at the individual level and makes predictions about future criminal justice involvement. The study uses historical data to train and evaluate the model as well as to evaluate the task performance of the MCRT clinicians. 

\paragraph{Explainable ML method:} The related use case for the task is Use Case No. 03 (deciding whether to intervene) and we use a feature attribution based post-hoc model agnostic explanation method for the evaluation (e.g., SHAP, LIME)

\paragraph{Inference Strategy:} In this trial we are interested in learning whether the explainable ML method is effective in helping the MCRT clinicians correctly pick individuals for mental health outreach. 
    At a minimum, the trial should evaluate the following hypotheses:
    \begin{enumerate}
        \item MCRT clinicians select individuals for outreach at a higher precision when presented with ML predictions than when they only have access to data
        \item MCRT clinicians select individuals for outreach at a higher precision when when they have access to explanations of ML predictions than when they have access only to ML predictions
    \end{enumerate}
    In order to evaluate these hypotheses, we need to create three experimental groups/arms where the clinicians have access to different levels of information: 
    \begin{enumerate}
        \item Clinicians have access only to the data of the individuals (Group1)
        \item Clinicians have access to the data, and the predicted risk score, e.g., a calibrated probability (Group2)
        \item Clinicians have access to the data, predicted risk score, and explanations generated by the post-hoc method for the prediction (Group3)
    \end{enumerate}
    In the ideal case, we would need a large enough user base where the unit of randomization could be the clinicians, and we could randomly assign clinicians to the three groups. However, in policy applications, having access to a domain expert user base of significant size to enable sufficient statistical power is rare. In that case, we could design the trial in stages like \cite{Jesus2021} did and randomize at the level of data instances (an individual at a specific point in time). While this limits the hypotheses we can evaluate (e.g., how clinicians with different levels of experience use and interact with ML explanations), we can still measure the efficacy of an explanation method in the use case.  
    With this setup, we can compare Group 2 against Group 1 to evaluate the first hypothesis and similarly compare Group 3 against Group 2 for the second.

\subsubsection{Importance of evaluating the performance-explainability trade-off (if any) for inherently interpretable models}
In addition to evaluating the effectiveness of ML explanations in helping domain experts in public policy settings, it is important to assess the viability of inherently interpretable models in terms of predictive performance.
As inherently interpretable models rely on carefully curated input features, exploring the trade-off between performance and scalability in practice is crucial to ascertain their broad applicability. 
To that end, the models should be implemented on several real policy problems, evaluating: 
    1) the trade-off between feature preparation efforts and performance, and
    2) their ability to generalize to future data under strong temporal dependencies.

The prospect of inherently interpretable ML models that are intelligible without sacrificing predictive performance certainly holds considerable appeal. However, their current implementations are limited to a handful of practical contexts. To understand potential trade-offs in practice, we must rigorously test these models against more opaque models across problem domains. Even if there are performance limitations, there may be critical applications where the intelligibility of the model cannot be compromised, and some applications where there could be built-in guardrails to protect against unintended harm. Understanding the limitations of methods through experimentation will help practitioners make more informed implementation decisions and support the complete spectrum of use cases.



\subsection{Gap 2: Existing methods are not explicitly designed for specific use-cases}
As discussed above, existing methods are developed with loosely defined or generic explainability goals (e.g., transparency) and without well-defined context-specific use-cases. 
As a result, methods are developed without understanding the specific requirements of a given domain, use-case, or user-base, resulting in a lack of adoption and sub-optimal outcomes. 

While several existing methods may be applicable for each use-case, their effectiveness in real-world applications is not yet well-established, meaning this potential applicability may fail to result in practical impact.
As more methods are rigorously evaluated in practical, applied settings as suggested above, gaps in their ability to meet the needs of these use-cases may become evident.
For instance, model agnostic methods such as LIME \cite{Ribeiro2016} and SHAP \cite{Lundberg2017} are capable of extracting input feature importance scores for individual predictions from otherwise opaque models.
However, it is unclear whether they can address needs such as generating explanations that are well-contextualized and truly interpretable by less technical users without sacrificing fidelity (e.g., to help a domain expert identify unreliable model predictions or an affected individual seek recourse).

\section{Conclusion}
Despite the development of a wide array of explainable ML methods, their efficacy in improving real-world decision-making systems is yet to be sufficiently explored. 
In this paper, we sought to characterize and understand this gap in the present literature in hopes that this effort can help structure future evaluations of these methods to better address their practical utility. 
First, we identified the primary set of use-cases for ML model explanations in the ML aided public policy decision making pipeline: 1) model debugging, 2) regulator trust \& model adoption, 3) deciding whether to intervene, 4) deciding how to intervene, and 5) recourse. 
For each use-case, we defined the goals of an ML explanation and the intended end-user.
Then, we summarized the existing approaches in explainable ML and identified the degree to which this work addresses the needs of the identified use-cases. 
We observed that, while the existing approaches are potentially applicable to the use-cases, their utility has not been thoroughly validated for any of the use-cases through well-designed empirical user studies.

Two main gaps were evident in the design and evaluation of existing work: 1) methods are not sufficiently evaluated on real-world contexts, and 2) they are not designed and developed with target use-cases and well-defined explainability goals in mind. 
In response to these gaps, we proposed several research directions to systematically evaluate the existing methods with problems with real policy goals, real-world data, and domain experts.

A key aim of this paper is to connect the ML research community who develop explainable ML methods to the problems and needs of the public policy and social good domains.
As computer scientists who develop and apply ML algorithms to social/policy problems in collaboration with government agencies and non-profits, we are ideally and uniquely positioned to understand both the existing body of work in explainable ML and the explainability needs of the domains such as public health, education, criminal justice, and economic development. 
Two main factors motivated us to compile this discussion: 1) despite the existence of a large body of methodological work in explainable ML, we failed to identify methods that we could directly apply to the problems we were tackling in the real world and 2) the frequent conversations initiated by our colleagues in the ML research community on how their methods could be better suited and developed for real-world ML problems.

We strongly believe explainable ML methods will prove to be a critical component of ML systems that are designed for policy and societal problems, where high-stakes decisions with significant impacts on people's lives create a moral imperative for these systems to perform well across all five use cases we discuss. 
As such, there is considerable potential for explainable ML to have a broad positive impact on society through these applications, but it will only have this impact if we design and develop these methods for explicitly defined use-cases and evaluate them in a way that demonstrates their effectiveness on those use-cases. 
Therefore, the goal of this paper was not to develop new algorithms, nor was it to conduct a thorough survey of explainable ML work (since there are already several excellent articles on that topic). 
Rather, our goal was to take the necessary first steps to bridge the gap between methodological work and real-world needs. 
We hope that this discussion will help the ML research community collaborate with the Policy and HCI communities to ensure that existing and newly proposed explainable ML methods are well-suited to meet the needs of the end-users to give them the confidence to implement and deploy them in systems that can benefit society.



\paragraph{Funding Statement}
This work was funded by the Block Center for Technology and Society at Carnegie Mellon University.

\paragraph{Competing Interests}
None

\paragraph{Ethical Standards}
The research meets all ethical guidelines, including adherence to the legal requirements of the study country.

\paragraph{Author Contributions}
Conceptualization: KA, KTR, HL, RG; 
Writing - Original Draft: KA, KTR; 
Writing - Review \& Editing: RG, HL; 
Investigation:  KA, KTR, HL; 
Supervision: RG; 
Funding Acquisition: RG; 
All authors approved the final submitted draft.

\paragraph{Data Availability}
Data sharing is not applicable to this manuscript as no new data were created or analyzed in this work.

\bibliographystyle{plainnat}
\bibliography{references}


\end{document}